\documentclass[11pt,a4paper]{article}
\usepackage[hyperref]{acl2021}
\usepackage{times}
\usepackage{latexsym}
\usepackage{graphicx}
\usepackage{tablefootnote}
\usepackage{subcaption}
\usepackage{amsmath}

\usepackage{microtype}

\aclfinalcopy % Uncomment this line for the final submission

\title{NLRG at SemEval-2021 Task 5: Toxic Spans Detection Leveraging BERT-based Token Classification and Span Prediction Techniques}

\author{Gunjan Chhablani\thanks{\hspace{2mm}Equal contribution. Author ordering determined by coin flip.} \\
  Dept. of CS\&IS\\
  BITS Pilani, Goa, India\\
 \texttt{\scriptsize chhablani.gunjan@gmail.com}\\
 
 \And
 
Abheesht Sharma\footnotemark[1] \\
  Dept. of CS\&IS\\
  BITS Pilani, Goa, India\\
 \texttt{\scriptsize f20171014@goa.bits-pilani.ac.in} \\

\AND

\textbf{Harshit Pandey}\footnotemark[1]  \\
  Dept. of CS \\
  Pune University, India \\
  \texttt{\scriptsize hp2pandey1@gmail.com} \\
  
\And
  Yash Bhartia  \\
  Dept. of CS\&IS\\
  BITS Pilani, Goa, India\\
  \texttt{\scriptsize f20190151@goa.bits-pilani.ac.in} \\
\And
 Shan Suthaharan \\
 Dept. of CS \\
  UNC-Greensboro, NC, USA\\
  \texttt{\scriptsize s\_suthah@uncg.edu}\\
}

\date{}

\begin{document}

\maketitle
\begin{abstract}
Toxicity detection of text has been a popular NLP task in the recent years. In SemEval-2021 Task-5 Toxic Spans Detection, the focus is on detecting toxic spans within English passages. Most state-of-the-art span detection approaches employ various techniques, each of which can be broadly classified into Token Classification or Span Prediction approaches. In our paper, we explore simple versions of both of these approaches and their performance on the task. Specifically, we use BERT-based models - BERT, RoBERTa, and SpanBERT for both approaches. We also combine these approaches and modify them to bring improvements for Toxic Spans prediction. To this end, we investigate results on four hybrid approaches - Multi-Span, Span+Token, LSTM-CRF, and a combination of predicted offsets using union/intersection. Additionally, we perform a thorough ablative analysis and analyze our observed results. Our best submission - a combination of SpanBERT Span Predictor and RoBERTa Token Classifier predictions - achieves an $F_{1}$ score of 0.6753 on the test set. Our best post-eval $F_{1}$ score is 0.6895 on intersection of predicted offsets from top-3 RoBERTa Token Classification checkpoints. These approaches improve the performance by 3\% on average than those of the shared baseline models - RNNSL and SpaCy NER.

\end{abstract}

\section{Introduction}
% Social media platforms provide an environment where people can learn about the trends and news, freely share their opinions and engage in discussions. Unfortunately, the lack of a moderating entity in these platforms has caused problems ranging from fake news to online harassment \cite{hosseini2017deceiving}. Methods such as manual moderation and crowd sourced filtering (upvotes/downvotes) are inefficacious and are not scale-able.
% and manual filtering is very time consuming, and and the fact that it can cause post-traumatic stress disorder-like symptoms to human annotators, 
% Hence, there have been many research efforts aimed at automating the process. 
% \cite{zampieri-etal-2019-semeval}. 
% One such crucial task is
Offensive language can include various categories such as threats, vilification, insults, calumniation, discrimination and swearing \cite{pavlopoulos-etal-2019-convai}. Detection of such language is necessary for ease of moderation of content on social media. Despite their popularity, toxicity detection tasks have focused majorly on sequence classification, rather than sequence tagging. Finding which spans make a comment or document toxic in nature is crucial in explaining the reasons behind their toxicity. Additionally, such attributions would allow for more efficient semi-automated quality-based moderation of content, especially for verbose documents, in comparison to quantitative toxicity scores.
% Although several toxicity (abusive language) detection datasets and models have been released, most of them classify whole comments or documents, and do not identify the spans that make a text toxic. Highlighting such toxic spans can assist human moderators (e.g., news portals moderators) who often deal with verbose comments, and who prefer attribution instead of just a system-generated unexplained toxicity score per post. The evaluation of systems that could accurately locate toxic spans within a text is thus a crucial step towards successful semi-automated moderation.

In SemEval-2021 Task-5, \citet{pav2020semeval} provide a dataset of 10k English texts filtered from  Civil Comments \cite{borkan2019nuanced} dataset. Each text is crowd-annotated with character offsets that make the text toxic. The task is to predict these character offsets given the text. The work presented in this paper aims to provide a comprehensive analysis of simple Token Classification (TC) and Span Prediction (SP) methods across multiple BERT-based models - BERT \cite{devlin-etal-2019-bert}, RoBERTa \cite{liu2019roberta} and SpanBERT \cite{joshi2020spanbert}. Additionally, we experiment with a few hybrid approaches - Multi-Span (MSP), where the model is trained on multiple spans simultaneously; Span+Token (SP-TC), where the model is trained on both kinds of tasks simultaneously; LSTM-CRF (LC), which uses a LSTM and CRF layer on top of BERT-based models; and a combination of predicted offsets for above techniques using union/intersection. In Section \ref{sec:background}, we perform a compendious literature survey. Section \ref{sec:models} elucidates our approach, including the modelling aspect, the various variants of the base model, and the different Hybrid Systems. In Section \ref{sec:experimental_setup}, we describe our experimental setup and hyperparameters used for our methods. Lastly, in Section \ref{sec:results_and_discussion} we analyze our results and perform ablative analysis on our systems.

\section{Background}
\label{sec:background}
Before the advent in research pertaining to toxic texts, \citet{warner-hirschberg-2012-detecting} modeled hate speech as a word sense disambiguation problem where SVM was used for classification of data. \citet{mehdad-tetreault-2016-characters} used RNN Language Model with character and token based methods to classify the text. Recently, however, toxic text detection has garnered a lot of attention \cite{10.1145/2872427.2883062, park-fung-2017-one, pavlopoulos-etal-2017-improved, wulczyn2017ex}.
The increase in offensive language research can partly be credited to various workshops such as Abusive Language Online\footnote{\url{https://sites.google.com/site/abusivelanguageworkshop2017/}} \cite{ws-2017-abusive} , as well as other fora, such as GermEval for German texts,\footnote{\url{https://projects.fzai.h-da.de/iggsa/}} or TRAC \cite{kumar-etal-2018-benchmarking} 
and Kaggle challenges\footnote{\href{https://www.kaggle.com/c/jigsaw-toxic-comment-classification-challenge}{Jigsaw Toxic Comment Classification Challenge}}.

\citet{Detoxify} introduced Detoxify, a comment detection library modeled using HuggingFace's transformers \cite{wolf-etal-2020-transformers} to identify inappropriate or harmful text online as a result of participation in three such challenges. In a contemporary work, \citet{pavlopoulos-etal-2020-toxicity} discuss context requirement for toxicity detection.

In SemEval 2020-Task 11 \cite{da-san-martino-etal-2020-semeval}, the first sub-task - Span Identification - aims at detecting the beginning and the end offset for the propaganda spans in news articles. This sub-task is similar to SemEval 2021-Task 5. The proposed approaches for the sub-task can be broadly classified into Span Prediction or Token Classification. Most teams use multi-granular transformer-based systems for token classification/sequence tagging \cite{khosla-etal-2020-ltiatcmu, morio-etal-2020-hitachi-semeval,patil-etal-2020-bpgc}. Inspired by \citet{souza2019portugese}, \citet{jurkiewicz-etal-2020-applicaai} use RoBERTa-CRF based systems. \citet{li-xiao-2020-syrapropa} use a variant of SpanBERT span prediction system.

\begin{figure*}
\begin{subfigure}[t]{0.30\textwidth}
  \centering
    \includegraphics[width=\textwidth]{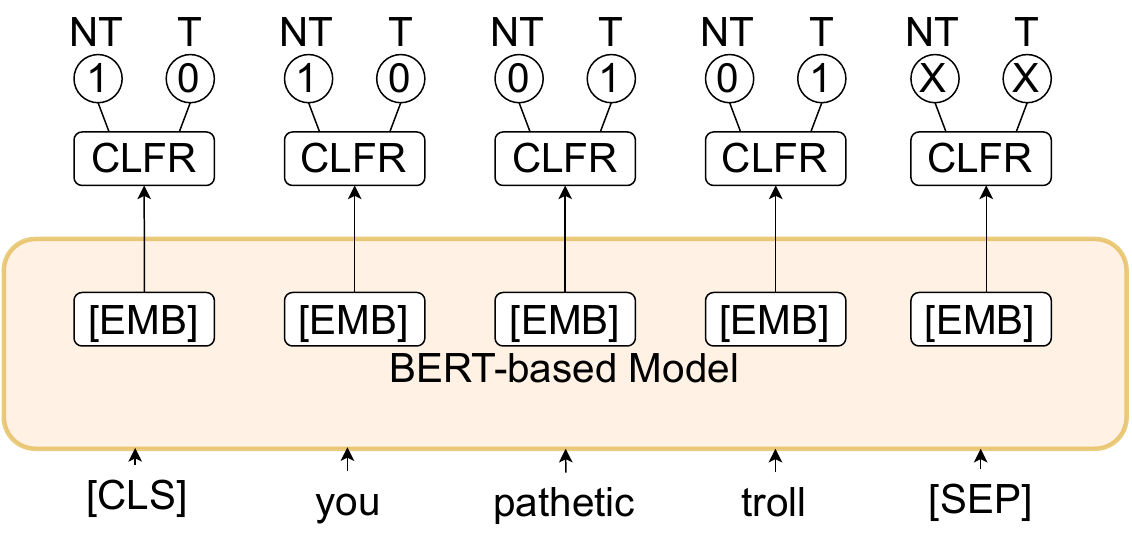}
      \caption{Token Classification}
    \label{fig:TC}
\end{subfigure}
\begin{subfigure}[t]{0.38\textwidth}
  \centering
    \includegraphics[width=\textwidth]{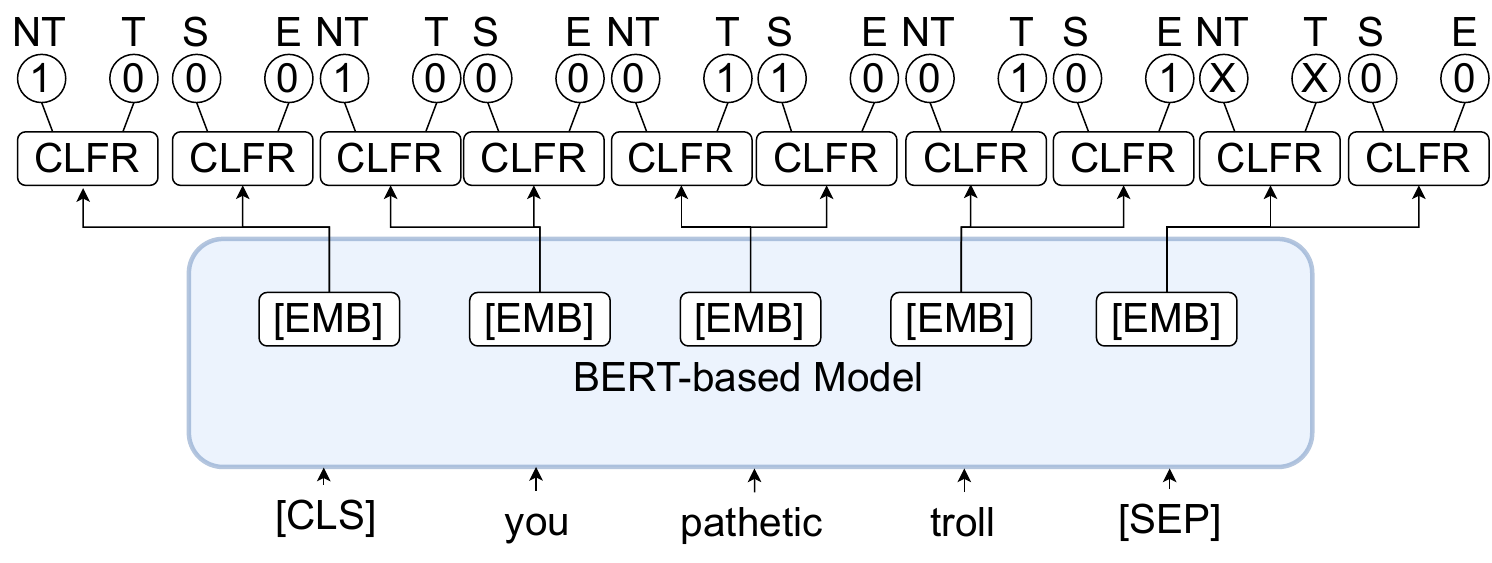}
      \caption{Span+Token}
    \label{fig:SP_TC}
\end{subfigure}
\begin{subfigure}[t]{0.30\textwidth}
  \centering
    \includegraphics[width=\textwidth]{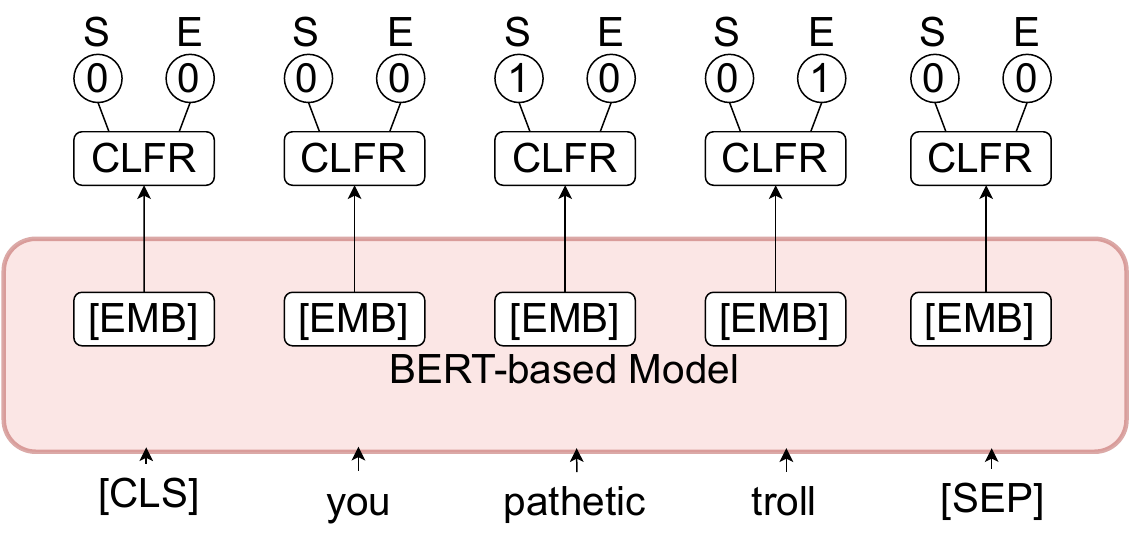}
      \caption{Span Prediction}
    \label{fig:SP}
\end{subfigure}
\begin{subfigure}[t]{0.33\textwidth}
  \centering
    \includegraphics[width=\textwidth]{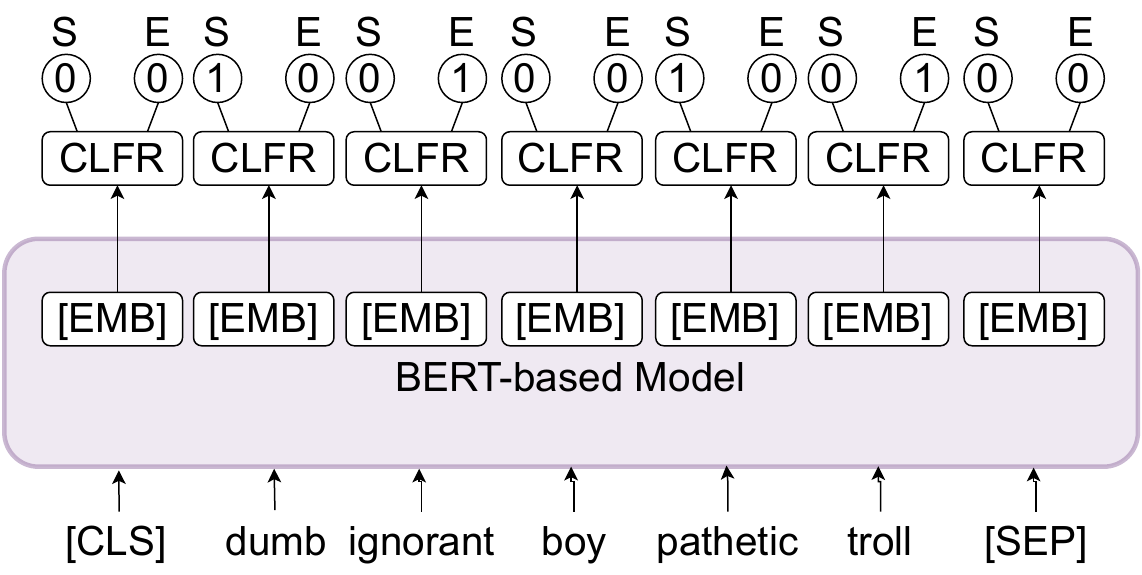}
      \caption{Multi-Spans}
    \label{fig:MSP}
\end{subfigure}\hfill
\begin{subfigure}[t]{0.33\textwidth}
  \centering
    \includegraphics[width=\textwidth]{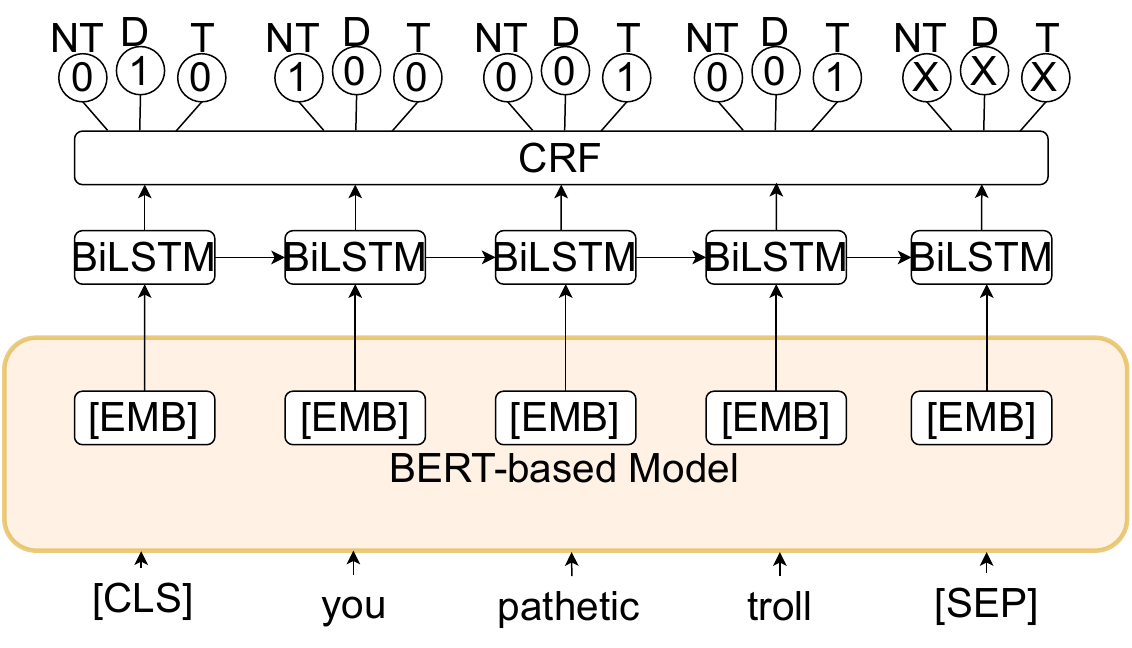}
      \caption{LSTM-CRF}
    \label{fig:LC}
\end{subfigure}

\caption{BERT-based Approaches\textsuperscript{*}}
\label{fig:BERT-based Approaches}
\tiny{\textsuperscript{*}CLFR = Classifier, [EMB] = Token Embedding, NT = Non-Toxic, T = Toxic, D = Dummy, X = Don't Care, S = Start, E = End.}

\end{figure*}

\section{Models}
\label{sec:models}
\subsection{Token Classification Models}
\subsubsection{Baseline Models}
From the models already provided with the dataset, we use RNNSL and SpaCy NER Tagging baselines for token-wise classification.

RNNSL model is a combination of a single Bi-LSTM layer with a randomly initialized embedding layer. It uses a three-label classification task for each word in the sentence. The labels used are: \textit{special token}, \textit{non-toxic word}, and \textit{toxic word}. For each word, the corresponding offsets are added to the predicted spans. A word with containing any toxic offset is marked as toxic during training.

SpaCy NER Tagging model is an NER classifier built on SpaCy Language Models. It is used to predict the entities which are labelled as \textit{TOXIC} in the text using the spans provided.

\subsubsection{BERT-based Token Classification Models}
These models comprise a BERT-based model and a classification layer over each final token embedding which predicts whether a token is toxic or not. Based on these classifications, we add the offsets for those tokens (not words) which are marked as toxic by the model. Figure \ref{fig:TC} represents a Token Classification Model.

\subsection{Span Prediction Models}
\label{sec:qa-models}
\subsubsection{BERT-based Span Prediction Models}
We use the BERT-based Span Prediction (Figure \ref{fig:SP}) models based on Extractive Question Answering systems similar to work on SQuAD \cite{rajpurkar-etal-2016-squad} and MRQA \cite{fisch-etal-2019-mrqa}. In these systems, the output at each token is a start logit and an end logit denoting whether that token is a start token or an end token of the span, depending on the softmax value. Since the Toxic Spans text can have multiple toxic spans, we take different contiguous spans from the given offsets, and make several `samples' out of the example. Each span becomes an `answer' for the particular text sample. We use the word \textit{`offense'} as a dummy question. Thus, each contiguous span leads to one `sample' for every example (Table~\ref{tab:qa-example}). 
\begin{table}[h]
\small
\centering
\begin{tabular}{p{0.6\linewidth}  p{0.3\linewidth}}
\hline
\textbf{Text} & \textbf{Spans}\\
\hline
%Agreed. I wouldn't call Trump %
...an idiot - just an embarrassingly uninformed, ignorant,... & idiot, ignorant\\
%inarticulate, bumbling nincompoop. POTUS, in its first 100 days, is fast on its way to losing all credibility.

\hline
\end{tabular}

\begin{tabular}{p{0.2\linewidth}p{0.5\linewidth}  p{0.1\linewidth}}

\textbf{Question} & \textbf{Context} & \textbf{Answer}\\
\hline
offense &
...an idiot - just an embarrassingly uninformed, ignorant,... & idiot\\
%Agreed. I wouldn't call Trump %
%inarticulate, bumbling nincompoop. POTUS, in its first 100 days, is fast on its way to losing all credibility.
\hline
offense &
...an idiot - just an embarrassingly uninformed, ignorant,... & ignorant\\
\hline
\end{tabular}
\caption{Conversion of Toxic Spans example to samples for single-span Span Prediction.}\label{tab:qa-example}
\end{table} 

We store the start index of the text, similar to the SQuAD \cite{rajpurkar-etal-2016-squad} dataset, and process the data to provide start and end token positions during training. The classifier layer on top of the encoder embeddings performs a binary classification task for start and end positions. A span is scored using the sum of predicted start and end logits. From top-K start and end logits, valid predicted answer spans\footnote{Valid spans are those which have end index greater than start index, and length less than a maximum span length.} are chosen during post-processing. A union of all the corresponding offsets is taken to give the final prediction for the example. A threshold is learned on the span scores using the resulting dev set $F_{1}$ score on offsets, which is then used for test set prediction. All spans with score above threshold are considered to be toxic spans.

% Add another tabular, with output from BERT after breaking the context.

\subsection{Hybrid Systems}
\subsubsection{Multi-Spans}
In Section \ref{sec:qa-models}, we allow each context to have multiple single-span answers during training. This is counter-intuitive, as the model is only trained to handle a single-span at a time, and expected to predict multiple single-spans during prediction. Two toxic spans in  text are equally important to predict, and thus, should not be shown at different times during training. To mitigate this issue, we try an approach which we refer to as the `Multi-Spans' (MSP) approach. Here, we take all the ground start and end token positions during training, and use Binary Cross Entropy on each of the start/end logits. This essentially treats the task as a multi-label classification problem. Hence, during training, all the ground spans are used in the same iteration with the example, and only one `sample' per example is generated. Figure \ref{fig:MSP} depicts a representation of the system. Note that two tokens - \textit{dumb} and \textit{pathetic} are marked as the start token. Similarly, both \textit{ignorant} and \textit{troll} are marked as the end token.

\subsubsection{LSTM-CRF}
A recently popular approach in Named-Entity Recognition tasks has been to use Conditional Random Fields (CRF) with BERT-based models. Inspired by the CRF-based approaches \cite{souza2019portugese,jurkiewicz-etal-2020-applicaai}, we use BERT-based models with a single BiLSTM layer and a CRF layer. During training, the CRF loss is used and during prediction, Viterbi Decoding is performed. Though CRF is generally used for word-level classification, we do not mask inner and end tokens for a word as it degrades dev set performance for our systems. Hence, all the tokens of a word are considered for classification.

\subsubsection{Spans+Token}
For this system, we use a combination of the two tasks - Token Classification and single-span Span Prediction. We use two classification layers on the token-wise embeddings - one for start and end prediction, and the other for token classification. Training is done simultaneously on both tasks, and the cross-entropy loss for each classifier is weighted. The overall loss is given as:
\begin{equation*}
    \begin{aligned}
        L(\hat{s},\hat{e},\hat{p},s,e,p) = -\sum_{t}\hat{p_{t}}\log{p_{t}}\\ - \frac{(\sum_{t}\hat{s_{t}}\log{s_{t}} + \sum_{t}\hat{e_{t}}\log{e_{t}})}{2}
    \end{aligned}
\end{equation*}
where $s_{t}$,$e_{t}$, and $p_{t}$ are labels for start, end and token classifiers for token $t$, while $\hat{s_{t}}$,$\hat{e_{t}}$ and $\hat{p_{t}}$ are predictions.
This is done to equally scale both SP and TC task losses.
During prediction, we consider top-K start and end scores. From the valid spans, the score is calculated as the average of start and end logit scores, as well as the mean of toxicity logits over the span under consideration. The score is given as:
\begin{equation*}
\begin{aligned}
    S(i_{s},i_{e}) = \frac{\hat{s}_{i_{s}}+\hat{e}_{i_{e}}}{2} + \frac{\sum_{k=i_{s}}^{i_{e}}\hat{t}_{k}}{e-s+1}
\end{aligned}
\end{equation*}
where $i_{s}$ and $i_{e}$ are start and end indices, $\hat{s}_{i_{s}}$ and $\hat{e}_{i_{e}}$ are start and end logits at those indices, and $\hat{t}_{k}$ is toxicity logit at index $k$.
A threshold, similar to Section \ref{sec:qa-models} is tuned on the dev set. The predicted offsets taken from the predicted spans are considered to be toxic.

\subsubsection{Combination of Offset Predictions}
\citet{chen2017checkpoint} proposed using the predictions from top few checkpoints and averaging the results to achieve better classification scores. Based on a similar line of thought, we also combine the predicted spans for various checkpoints of a model, as well as across different models using union or intersection.

\section[Experimental Setup]{Experimental Setup\footnote{Our code can be found at: \url{https://github.com/gchhablani/toxic-spans-detection}.}$^{,}$\footnote{We also use Integrated Gradients to understand what the models focus on. For discussion, see Appendix \ref{appendix: IG}.}}
\label{sec:experimental_setup}
\subsection{Hardware Requirements}
The training and the evaluation of systems was performed on Google Colab's free GPU (NVIDIA K80/P100). The training time varies with the models. For each model, it is around 4-6 hours, which is well-within the 12 hour limit of Colab.
\subsection{Models \& Hyperparameters}
\label{subsubsec:models_hparams}
For RNNSL, a Keras-based BiLSTM model is provided. We use a max length of 192, batch size of 32 and a dropout of 0.1. The training is done using Adam Optimizer with early stopping ($patience\_period=3$), which in our case halts at 5 epochs. The embedding/hidden\_state size used is 200. A threshold is used to classify a word as toxic on the predicted toxic word probability. This threshold is tuned on the trial dataset. For SpaCy, the \textit{en\_core\_web\_sm} model is used with 30 iterations.

For all BERT-based models, we use HuggingFace's transformers \cite{wolf-etal-2020-transformers} in PyTorch. For CRF, we use the pytorch-crf \cite{pytorch-crf} library. We use a batch size of 4, train for 3 epochs,  use a linear learning rate decay, and an AdamW optimizer with a weight decay of 0.01. The initial learning rate is $2\mathrm{e}{-5}$. During tokenization, the maximum length allowed is 384, with the exception of RoBERTa Span+Token where it is 512. We use \textit{LARGE} models for all - BERT, RoBERTa and SpanBERT, unless otherwise specified.

For Token Classification, we add a label for the \textit{[CLS]} token if the percentage of toxic offsets in text is greater than 30\% in order to provide a proxy text classification objective for the system. For span-based models, the K used for top-K start and top-K end logit selection is 20, and the maximum allowed answer length is 30 tokens. For LSTM-CRF systems, a dummy label is used for the \textit{[CLS]} token, while the prediction mask for other special tokens is set to 0. A dropout of 0.2 is used. For Span Prediction systems, the overlapping stride is set to 128.

The training dataset used is \textit{tsd\_train.csv} and the dev set used is \textit{tsd\_trial.csv} file, unless otherwise specified. For all systems, we evaluate the $F_{1}$ scores using the provided script on the checkpoints which give the lowest dev set loss.

\section{Results and Analysis}
In favor of brevity, for this section, we use the following abbreviations: {BT=BERT, RBTa=RoBERTa, SBT=SpanBERT, SP=Span Prediction, TC=Token Classification, MSP=Multi-Span, LC=LSTM-CRF, B=Base, TBT=ToxicBERT, TRBTa=ToxicRoBERTa, TT=Trained on Train+Trial, (x,$\cap$)=Intersection of offsets from x-best checkpoints,  (x,$\cup$)=Union of offsets from x-best checkpoints.}
\label{sec:results_and_discussion}
\begin{table}[!htb]
\small
    \centering
    \begin{tabular}{|c|c|c|c|}
    \hline
    \textbf{Model} & \textbf{Train $F_{1}$} & \textbf{Trial $F_{1}$} & \textbf{Test $F_{1}$} \\\hline\hline
    RNNSL     & 0.5904      & 0.5904      & 0.6514      \\\hline
    SpaCy     & 0.6282      & 0.5729      & 0.6573      \\\hline\hline
    BT-TC      & 0.6944      & 0.6942      & 0.6781      \\\hline
    RBTa-TC    & 0.6791      & 0.6769      & 0.6834      \\\hline
    SBT-TC     & 0.6873      & 0.6789      & \textbf{0.6856}      \\\hline\hline
    BT-SP      &  0.6639           &    0.6465         &        0.6663     \\\hline
    RBTa-SP    & 0.6401      & 0.6386      & 0.6665      \\\hline
    SBT-SP     & 0.6432      & 0.6212      & 0.6561      \\\hline\hline
    BT-MSP     &  0.5218           &    0.4941         &   0.5406          \\\hline
    RBTa-MSP   & 0.5056      & 0.4886      & 0.5244      \\\hline
    SBT-MSP    & 0.5190      & 0.5004      & 0.5084      \\\hline\hline
    BT-SP-TC   &   0.6676          &     0.6214        & 0.6186            \\\hline
    RBTa-SP-TC & 0.6395            &   0.6101          &     0.5901        \\\hline
    SBT-SP-TC  &      0.6608       &   0.6491          &     0.5959        \\\hline\hline
    BT-LC      &  0.6887           &    0.6843         &   0.6835          \\\hline
    RBTa-LC & 0.7236 &  0.6861 &   0.6787       \\\hline
    SBT-LC     &     0.7200        &     0.6982        &    0.6801             \\\hline
    \end{tabular}
    \caption{$F_{1}$ scores for our approaches (Post-Eval).} 
    \label{tab:post_eval_scores}
\end{table}

In Table \ref{tab:post_eval_scores}, we mention scores for our approaches. The scores are evaluated are performed after the evaluation phase, using the hyperparameters mentioned in Section \ref{subsubsec:models_hparams}. We observe that the highest score is obtained by SBT-TC (0.6856). The baseline scores (RNNSL/SpaCy) are good ($\approx$0.65) considering that these models are not pre-trained. Notably, SP systems perform worse than their TC counterparts. A good reason could be the self-attention used in BERT-based models. Since the interaction is between tokens, and not spans, it is expected that each token is well represented and less consideration will be given to the span representation around a single token. The reason why SBT-TC performs best out of all the LARGE models could be the random-spans Masked Language Modeling used in its pre-training. However, BERT and RoBERTa take over for other approaches.\\ 
LSTM-CRF approaches perform as good as Token Classification approaches, and BT-LC achieves the second highest score (0.6835). MSP performs poorly, in contrast to what is expected. Multi-Span Extraction is still an active problem in Deep NLP with only a few recent works \cite{segal-etal-2020-simple, yang2020multispan} on it, which still incorporate sequence tagging approaches. Spans+Token approaches perform better than Multi-Span, but are worse than both TC and SP approaches across all BERT-based models. \\
\begin{table}[!htb]
\centering
\begin{tabular}{|c|c|}
\hline
\textbf{Combination} & \textbf{Test $F_{1}$}\\\hline
RBTa-TC(3,$\cup$)   & 0.6765 \\\hline
RBTa-TC(3,$\cap$)  & \textbf{0.6895}  \\\hline
SBT-SP(3,$\cup$) & 0.5879  \\\hline
SBT-SP(3,$\cap$)  & 0.6585  \\\hline
RBTa-TC(3,$\cup$)$\cup$SBT-SP & 0.6573\\\hline
RBTa-TC(3,$\cup$)$\cap$SBT-SP & 0.6765 \\\hline
RBTa-TC$\cup$ SBT-SP(3,$\cup$)& 0.5840\\\hline
RBTa-TC$\cap$SBT-SP(3,$\cup$) & 0.6883\\\hline
\end{tabular}
\caption{$F_{1}$ scores for combined predictions.}
\label{tab:combined_scores}
\end{table}
Lastly, from combined checkpoint predictions (Table \ref{tab:combined_scores}), we get out best scoring system - RBTa-TC(3,$\cap$) - which achieves a score of 0.6895. However, our best official submission\footnote{The most significant of our official submission scores are present in Appendix \ref{Appendix A}.} was a variant of the third best combination - RBTa-TC(3,$\cup$)$\cap$SBT-SP (0.6765). It is also observed that intersection approaches perform better than corresponding union and single checkpoints approaches, while union approaches perform worse than single checkpoints. This means that the individual checkpoints are predicting some extra offsets to be toxic.

\subsection{Ablative Analysis}

\begin{table}[!htb]
\small
    \centering
    \begin{tabular}{|c|c|c|c|}
    \hline
     \textbf{Model} & \textbf{Train $F_{1}$} & \textbf{Trial $F_{1}$} & \textbf{Test $F_{1}$}\\
    \hline
    TBT-TC & 0.6753	& 0.6628 &	0.6792\\\hline
    TRBTa-TC &0.7244 &	0.6954&	0.6773\\\hline
    TBT-SP &0.6638	&0.6560	&0.6584\\\hline
    TRBTa-SP &0.6475	&0.6358	&0.6746\\\hline
    \hline
    BT-B-TC & 0.6966	& 0.6746 &	\textbf{0.6881}\\\hline
    RBTa-B-TC &0.6641 &	0.6482&	0.6834\\\hline
    BT-B-SP &0.6605	&0.6434	&0.6611\\\hline
    RBTa-B-SP &0.6481	&0.6464	&0.6661\\\hline
    \hline
    RNNSL-TT& 0.6844	& 0.6882 &	0.6259\\\hline
    RBTa-TC-TT &0.7707 &	0.7788&	0.6823\\\hline
    SBT-SP-TT &	0.7116 &0.7092	&0.6669\\\hline
    \end{tabular}
    \caption{$F_{1}$ scores for ablative approaches.\footnotemark[8]}
    \label{tab:ablative_scores}
\end{table}
\label{sec:ablative_analysis}
In Table \ref{tab:ablative_scores}, we present results on TBT\footnote{\url{https://huggingface.co/unitary/toxic-bert}} and TRBTa\footnote{\url{https://huggingface.co/unitary/unbiased-toxic-roberta}} for TC and SP approaches. These are \textit{BASE} models fine-tuned on the Civil Comments Dataset. Since the Toxic Spans dataset has similar text data, we expect these models to perform better than \textit{BASE} models. We observe that TBT-TC and TRBTa-SP perform slightly better than BT-TC and RBTa-SP, despite being \textit{BASE} models. Also, BT-SP and RBTa-TC are only slightly better than their `Toxic counterparts.
\\Yet, in comparison, \textit{BASE} models - BT-B and RBTa-B, without any multi-stage pre-training perform better than their `Toxic' counterparts, and are comparable, if not better than their \textit{LARGE} counterparts. This means that there not enough data for \textit{LARGE} models, and hence, they tend to overfit. However, the reasons behind worse performance of `Toxic' systems is unclear.
\\We also evaluate scores for a few systems on the test set after 3 epochs of training on both train and trial data (-TT). We observe that the performance on both train and trial datasets increases significantly ($\approx$7-10\%), showing that these datasets have similar distribution. However, the performance on test decreases for RBTa-TC-TT and RNNSL-TT in comparison to the Table \ref{tab:post_eval_scores}, which shows that test set distribution might be slightly different for TC task. For SBT-SP-TT, we see a slight increase, showing scope of improvement for SP systems with more data.\\
Lastly, we evaluate the token-based predictions and span-based predictions for SBT SP-TC separately. Surprisingly, token predictions achieve a $F_{1}$ score of 0.6522 on the test set, which is much better than using both token and spans (0.5959). However, for span-based predictions, we only achieve an $F_{1}$ score of 0.1510. This means that the system is focusing heavily on token-based-predictions. Hence, we need to re-evaluate our architectural decisions in order to successfully incorporate both token and spans together.

% \begin{table}[]
%     \centering
%     \begin{tabular}{|c|c|c|c|}
%     \hline
%     \textbf{Model} & \textbf{Train $F_{1}$} & \textbf{Trial $F_{1}$} & \textbf{Test $F_{1}$} \\\hline
%     ToxicBT-TC & 0.6753	& 0.6628 &	0.6792\\\hline
%     ToxicRBTa-TC &0.7244 &	0.6954&	0.6773\\\hline
%     ToxicBT-SP &0.6638	&0.6560	&0.6584\\\hline
%     ToxicRBTa-SP &0.6475	&0.6358	&0.6746\\\hline
%     \hline
%     BT-B-TC & 0.6966	& 0.6746 &	\textbf{0.6881}\\\hline
%     RBTa-B-TC &0.6641 &	0.6482&	0.6834\\\hline
%     BT-B-SP &0.6605	&0.6434	&0.6611\\\hline
%     RBTa-B-SP &0.6481	&0.6464	&0.6661\\\hline
%     \hline
%     RNNSL-TT& 0.6966	& 0.6746 &	0.6881\\\hline
%     RBTa-TC-TT &0.6641 &	0.6482&	0.6834\\\hline
%     SBT-SP-TT &	&	&\\\hline
%     \end{tabular}
%     \caption{$F_{1}$ scores for Ablation Approaches.\footnotemark[1]}
%     \label{tab:ablation_eval_scores}
% \end{table}

\section{Conclusion}
Based on our results and analysis, we conclude that Token Classification systems have an edge over Span Prediction methods on this task. \textit{BASE} models perform better than \textit{LARGE} models in either of the approaches, which could imply need for more data to train \textit{LARGE} models. Our Multi-Span approach performs poorly, but Span+Token approach shows some promise and we need to re-evaluate our architectural choices. The reason why ToxicBERT/ToxicRoBERTa perform worse than \textit{BASE} models is also an avenue for further analysis. Finally, our individual BERT-based models tend to predict extra offsets for the task. While checkpoint ensembling using intersection is a good way to address this issue, we will explore other remedies in a future work.

\section*{Acknowledgments}
We would like to acknowledge the help and moral support provided to us by Rajaswa Patil\footnote{ \url {https://rajaswa.github.io/} } and Somesh Singh\footnote{\url {https://someshsingh22.github.io/} }. We would also like to express our gratitude to our colleagues at the Language Research Group (LRG)\footnote{\url{https://lrg.saidl.in/}}, who have been with us at every stepping stone.

\bibliographystyle{acl_natbib}
\bibliography{anthology,acl2021}

\appendix
\section{Official Submissions}
\label{Appendix A}
During the evaluation period, we performed a `cleaning' of the data by removing starting/trailing whitespace and punctuation characters in spans. Additionally, we include those partial words in spans which had more than half the number of characters in the span, and discard remaining partial words from spans. We considered this version of the \textit{tsd\_train.csv} and \textit{tsd\_trial.csv} to be `clean train' and `clean trial', respectively. During the post-eval period, we found out potential issues with the cleaning, and thus, we use original files. Additionally, since the distribution of \textit{tsd\_test.csv} is expected to be similar to \textit{tsd\_train.csv} and \textit{tsd\_trial.csv}, the scores are much better for models trained on \textit{tsd\_train.csv} file instead of \textit{clean\_train.csv}. However, some of our official submissions were from systems trained on the `clean train' data. Keeping that in mind, we report our official scores for our top-few approaches in Table \ref{tab:official_submission_scores}.
\begin{table}[h]
\small
    \centering
    \begin{tabular}{|p{0.4\linewidth}|p{0.3\linewidth}|p{0.15\linewidth}|}
    \hline
    \textbf{Model}&\textbf{Trained On}&\textbf{Test $F_{1}$}\\
    \hline
    RNNSL & Train+Trial & 0.6446\\\hline
    SpaCy & Train+Trial & 0.6470\\\hline
    RNNSL $\cup$ Spacy & Train+Trail & 0.6510\\\hline
    RBTa-TC&Clean Train&0.6270\\\hline
    RBTa-TC(3,$\cup$)&Clean Train&0.6469\\\hline
    SBT-SP&Train&0.6631\\\hline
    RBTa-TC(3,$\cup$) $\cap$ \newline SBT-SP&Clean Train, Train&\textbf{0.6753}\\
    \hline
    
    \end{tabular}
    \caption{Official Submission Scores}
    \label{tab:official_submission_scores}
\end{table}

\section{Integrated Gradients}
\label{appendix: IG}
We use Integrated Gradients\cite{sundarajan2017axiomatic} from the Captum\cite{kokhlikyan2020captum} library for qualitative analysis of predictions for the SpanBERT-SP, and the RoBERTa-TC models. We calculate Integrated Gradients of the targets with respect to the embedding layer outputs. The Riemann Right numerical approximation method is used, with \textit{n\_steps=50}.  Following \citet{ramnath-etal-2020-towards}, we calculate token-wise importance distributions and word-wise distributions for a few examples. We refer the paper to the reader for more details. 

For the Token Classification model, the targets are softmax outputs of toxicity logits of those tokens which the model predicts to be toxic, with a score greater than 0.5. For all such toxicity logits as targets, we calculate attributions with respect to the embedding layer outputs for all the tokens, and average them to get token-wise importance scores. For the Span Prediction model, we find start and end indices for all the predicted spans, and calculate respective attributions, add them, and then average them to get token-wise importance scores.

\begin{figure}[ht]
\begin{subfigure}[t]{0.49\textwidth}
  \centering
    \includegraphics[width=\textwidth]{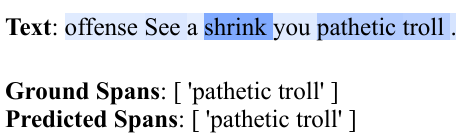}
      \caption{SpanBERT Span Prediction}
    \label{fig:SBT_IG}
\end{subfigure}
\begin{subfigure}[t]{0.49\textwidth}
  \centering
    \includegraphics[width=\textwidth]{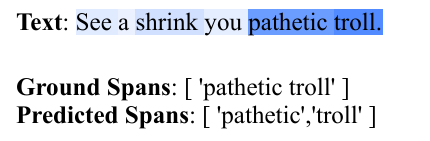}
    \caption{RoBERTa Token Classification}
    \label{fig:RBTa_IG}
\end{subfigure}
\caption{Qualitative Example of Attributions - Example 1}
\label{fig:ig_1}
\end{figure}

\begin{figure}[ht]
\begin{subfigure}[t]{0.49\textwidth}
  \centering
    \includegraphics[width=\textwidth]{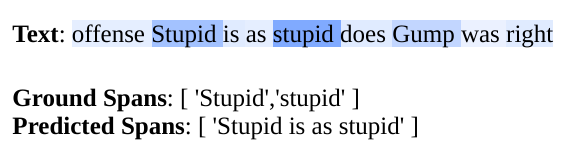}
      \caption{SpanBERT Span Prediction}
    \label{fig:SBT_IG_2}
\end{subfigure}
\begin{subfigure}[t]{0.49\textwidth}
  \centering
    \includegraphics[width=\textwidth]{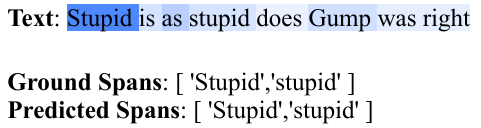}
    \caption{RoBERTa Token Classification}
    \label{fig:RBTa_IG_2}
\end{subfigure}
\caption{Qualitative Example of Attributions - Example 2}
\label{fig:ig_2}
\end{figure}

\begin{table*}[p]
\centering
\begin{tabular}{p{0.95\textwidth}}
\textbf{Text} : Why does this author think she can demand, or is owed anything from either of these two people? One guy is a goon, the other is illiterate. They aren't law makers, teachers, or in any kind moral authority position. They are entertainers who get punched for her pleasure, and will likely live out their days mentally debilitated from the repeated blows to the head.\\

Do we get to comb deeply through this authors personal history and determine all the groups she owes apologies or explanations to? Why not? As an opinion maker in a national news paper and instructor of young people, she has far, far more influence on Canadians than two ignorant punchies.\\

The arrogance of these pseudo-intellectual academics is astounding. Since they are so enlightened and pure, YOU owe THEM an explanation and an apology as to why you're so dumb and ignorant.\\
\textbf{Ground Spans}:  [dumb] 
\end{tabular}
\begin{tabular}{|p{0.3\textwidth}|p{0.55\textwidth}|}
\hline
BT-B-SP & [] \\\hline
BT-B-TC & [dumb, ignorant] \\\hline
BT-LC & [dumb, ignorant] \\\hline
BT-MSP & [dumb] \\\hline
BT-SP & [] \\\hline
BT-TC & [dumb, ignorant] \\\hline
BT-SP-TC & [dumb and ignorant] \\\hline
RBTa-TC(3,$\cap$) & [dumb, ignorant] \\\hline
RBTa-TC$\cap$SBT-SP(3,$\cup$) & [dumb, ignorant] \\\hline
SBT-SP(3,$\cap$) & [] \\\hline
RBTa-TC(3,$\cup$)$\cap$SBT-SP & [] \\\hline
RBTa-TC(3,$\cup$) & [go, dumb, ignorant] \\\hline
RBTa-TC$\cup$ SBT-SP(3,$\cup$) & [dumb and ignorant] \\\hline
SBT-SP(3,$\cup$) & [dumb and ignorant] \\\hline
RBTa-TC(3,$\cup$)$\cup$SBT-SP & [go, dumb, ignorant] \\\hline
RNNSL & [ignorant, dumb, ignorant] \\\hline
RNNSL-TT & [goon, ignorant, dumb, ignorant] \\\hline
RBTa-B-SP & [] \\\hline
RBTa-B-TC & [dumb] \\\hline
RBTa-LC & [on, ignorant, dumb, ignorant] \\\hline
RBTa-MSP & [] \\\hline
RBTa-SP & [] \\\hline
RBTa-TC & [dumb, ignorant] \\\hline
RBTa-SP-TC & [ignorant, dumb and ignorant] \\\hline
RBTa-TC-TT & [dumb, ignorant] \\\hline
SpaCy & [ignorant] \\\hline
SBT-LC & [ignorant, dumb, ignorant] \\\hline
SBT-MSP & [dumb and ignorant] \\\hline
SBT-SP & [] \\\hline
SBT-SP-TT & [dumb and ignorant] \\\hline
SBT-TC & [ignorant, dumb, ignorant] \\\hline
SBT-SP-TC & [ignorant, dumb and ignorant] \\\hline
TBT-SP & [] \\\hline
TBT-TC & [ignorant] \\\hline
TRBTa-SP & [] \\\hline
TRBTa-TC & [dumb, ignorant] \\\hline
\end{tabular}
\caption{The prediction output of the models for an example in the test set.}
\label{tab:model_predictions}
\end{table*}

We observe in Figure \ref{fig:SBT_IG} that the Span Prediction model performs correct prediction. However, on average, the word \textit{`shrink'} gets higher importance than \textit{`pathetic troll'}. This is in contrast with Figure \ref{fig:RBTa_IG} where the Token Detection model misses out on space (because it only considers tokens) and focuses more on the words \textit{`pathetic', `troll'}. However, the word `shrink' seems to be important in both cases. This means that while Token Classification models perform better, there are cases which are missed by these approaches. Additionally, some words outside of the span may contribute to toxicity of a particular span. We will be analyzing such words in a future work.

\section{Model Predictions}
\label{appendix: Model Predictions}
The predictions of the various systems for one example that is present in the test set, are listed in Table~\ref{tab:model_predictions}. The examples provide the following intuition about the data and the systems:
\begin{itemize}
    \item The spaces in between the words are, predictably, ignored by the the token based models. Moreover, the conjunctives like `and' are ignored as well. This means that additional post-processing of the data will lead to improvements in performance of token classification systems.
    \item Sometimes, random words like `go' and `on' are selected to be toxic, which means that these types of prepositions and verbs can be removed by exact matching in the string, unless they form parts of larger spans.
    \item The best checkpoints of the span-based models tend to predict empty spans for the selected example. However, when using checkpoint ensembling, we see that union models return accurate spans. %give out close to correct spans.
    \item The ground spans are not entirely correct and are ambiguous. For example, it is not clear whether the word `ignorant' should be considered to be toxic. The models, based on other examples, predict `ignorant` to be toxic, but it is not present in the ground spans. This means that finding the toxic spans is not a trivial task for humans, and annotation can not be performed easily by crowd-workers. 
    \item In some cases, one of the occurrences of the word `ignorant' is considered to be toxic, while the other is predicted to be benign. The first instance of `ignorant' does not seem to be as toxic as the second instance and therefore, more analysis needs to be done to determine the `degree' of toxicity of the spans. This can be a good direction for future research.
\end{itemize}

\end{document}